\documentclass[conference]{IEEEtran}
\IEEEoverridecommandlockouts
\usepackage{url}
\usepackage[hidelinks]{hyperref}
\usepackage{cite}
\usepackage{amsmath,amssymb,amsfonts}
\usepackage{algorithmic}
\usepackage{graphicx}
\usepackage{booktabs}
\usepackage{textcomp}
\usepackage{xcolor}
\def\BibTeX{{\rm B\kern-.05em{\sc i\kern-.025em b}\kern-.08em
    T\kern-.1667em\lower.7ex\hbox{E}\kern-.125emX}}
\begin{document}

\title{Data-Efficient Learning from Demonstration for Humanoid Robots via Feature-Level Perception and Prompt-Based Object Detection}

\author{
\IEEEauthorblockN{
Keyhan Rayati\IEEEauthorrefmark{1},
AliReza Beigy\IEEEauthorrefmark{1},
Anoosheh Saadati\IEEEauthorrefmark{1},
Pourya Shahverdi\IEEEauthorrefmark{2},\\
Mehdi Tale Masouleh\IEEEauthorrefmark{1},
Ahmad Kalhor\IEEEauthorrefmark{1},
Wing-Yue Geoffrey Louie\IEEEauthorrefmark{2}
}
\IEEEauthorblockA{
\IEEEauthorrefmark{1}
Human and Robot Interaction Lab, University of Tehran, Tehran, Iran\\
Emails: keyhan.rayati@ut.ac.ir, alireza.beigy@ut.ac.ir, anoosheh.saadati@ut.ac.ir,\\
m.t.masouleh@ut.ac.ir, ahmad.kalhor@ut.ac.ir
}
\IEEEauthorblockA{
\IEEEauthorrefmark{2}
Intelligent Robotics Laboratory, Oakland University, Rochester, Michigan, USA\\
Emails: pourya.shahverdi@oakland.edu, geoff.louie@oakland.edu
}
}

\maketitle

\begin{abstract}
Learning from Demonstration (LfD) allows robots to learn new skills from human examples; however, many vision-based methods require large datasets and resource-intensive end-to-end training.
This paper introduces a data-efficient LfD framework for humanoid robots that learns manipulation skills from compact, task-specific features instead of raw images.
The proposed approach integrates layered visual preprocessing, Kinect-based human joint estimation, prompt-based object detection, 
and depth information to extract human motion features and three-dimensional object locations from demonstration videos.
These features are used to train a temporal deep learning model that predicts robot joint-angle trajectories for execution.
The framework is implemented on a NAO humanoid robot, evaluated quantitatively in simulation, and demonstrated on physical hardware.
In simulation, the robot achieves a 78\% success rate on a pick-and-place task using only 16 demonstrations, which increases to 86\% when trained with 100 demonstrations.
The framework also achieves a 71\% success rate on a pushing task using eight demonstrations, highlighting its effectiveness for data-efficient humanoid skill acquisition.
\end{abstract}


\section{Introduction}
\label{sec:introduction}

Learning from Demonstration (LfD) offers an effective way for robots to acquire new skills through the observation of human actions, rather than depending on explicitly programmed control policies.
This is especially beneficial for humanoid robots, as creating task-specific controllers for manipulation is labor-intensive and difficult to generalize across different objects, 
environments, and users.
Nonetheless, acquiring dependable humanoid manipulation skills from a limited number of demonstrations remains difficult due to perceptual noise, 
embodiment differences between humans and robots, and the restricted computational resources available on several humanoid platforms.

LfD has been extensively studied for transferring human skills to robots, from traditional trajectory-based techniques such as Dynamic Movement Primitives
 to recent data-driven methods for robot skill learning from demonstrations~\cite{r14,saveriano2023dynamic,r27}.
Vision-based robot learning has also shown strong performance in manipulation, from spatial correspondence models such as Transporter Networks~\cite{r13}
 to large-scale visuomotor policies such as RT-1~\cite{r20}.
However, many of these methods still rely on raw visual inputs, large demonstration datasets, or considerable task-specific adjustment~\cite{r13,r18,r20,r21}.
For humanoid robots, these challenges become harder because of embodiment differences, restricted joint limits, and the need to translate human motion and task
 structure into robot-executable trajectories~\cite{r24,dariush2009online,r6}.
Recent studies have explored more data-efficient learning from human videos and robot-free demonstrations, including keypoint-based representations, one-shot
 transfer, humanoid manipulation from human play videos, whole-body control from monocular videos, 
 and mobile manipulation from robot-free demonstrations~\cite{r11,demoDiffusion2025,mimicDroid2025,hdmi2025,hommi2026}.
Still, learning robust humanoid manipulation skills from only a few task demonstrations remains difficult, especially when the perception and policy learning
 stages are trained directly from visual data.
This motivates the feature-level approach used in this work, where prompt-based object detection, depth information, and skeleton-based motion features are
 extracted before training the robot trajectory model.

This research introduces a data-efficient LfD framework for humanoid robots that acquires manipulation skills using compact state representations rather than raw images.
The proposed approach extracts task-relevant information from human demonstration videos, including human joint positions, prompt-detected objects, 
and depth-based three-dimensional object locations.
Prompt-based object detection localizes task objects without requiring a task-specific detector, while skeleton tracking provides human motion data for estimating robot joint trajectories.
The collected features are then used to train a task-specific temporal model that predicts joint-angle commands for execution on a NAO humanoid robot.

The framework is assessed quantitatively in simulation and demonstrated on real hardware, including upper-body imitation and two manipulation tasks: pick-and-place and object pushing.
In the simulated pick-and-place task, the robot attains a success rate of 78\% with 16 demonstrations and 86\% with 100 demonstrations.
The framework also attains a 71\% success rate in the pushing task with only eight demonstrations.
These results suggest that feature-level perception can facilitate humanoid skill acquisition with limited demonstration resources.

The primary contributions of this work are summarized as follows:
\begin{itemize}
    \item A data-efficient LfD framework is provided for humanoid manipulation using compact state representations instead of raw image-based policy learning.
    \item A layered perception pipeline is established to integrate Kinect-based skeleton tracking, prompt-based object detection, 
    and depth data for extracting task-relevant demonstration features.
    \item A temporal prediction model with explicit step and task-state embeddings is used to condition progress during autoregressive joint-trajectory rollout.
    \item The framework is evaluated on a NAO humanoid robot through quantitative simulation experiments and representative real-world demonstrations, showing upper-body imitation, 
    pick-and-place, and pushing actions with limited demonstration data.
\end{itemize}

\noindent \textbf{Supplementary Material:}
A video demonstration of the proposed framework is available online~\href{https://youtu.be/cuUF_SJkuuI}{[Video]}.
The code implementation is available online~\href{https://github.com/TaarLab/NaoLfD}{[Code]}.

\section{Proposed Framework}
\label{sec:method}

\begin{figure*}[t]
\centering
\includegraphics[width=\linewidth]{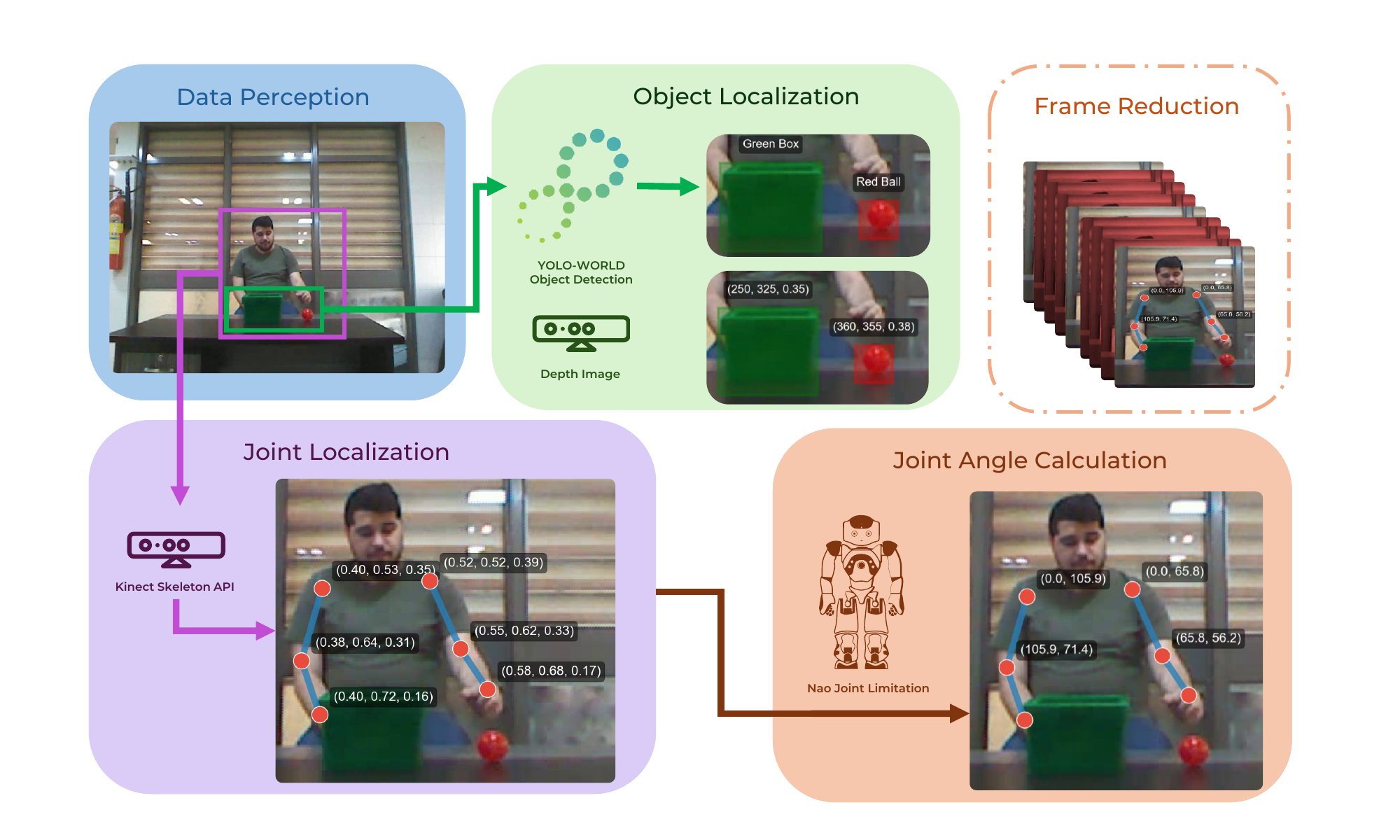}
\caption{Overview of the proposed LfD framework, including demonstration capture, feature extraction, temporal modeling, and robot execution.}
\label{fig:framework_overview}
\end{figure*}

The proposed framework learns robot arm trajectories from human demonstration videos using a compact feature representation rather than raw image sequences.
As illustrated in Fig.~\ref{fig:framework_overview}, the pipeline consists of demonstration capture, feature extraction, temporal model training, and robot execution.
RGB images, depth information, and human skeletal joint positions are captured using a Kinect camera as a person performs the task.
Task-relevant objects are localized using YOLO-World prompt-based object detection~\cite{cheng2024yoloworld}, with categories such as ``red ball'', ``green box'', or ``block'' provided as textual prompts.
The detected image-space coordinates are matched with Kinect depth data and the camera model to estimate three-dimensional object positions.
Human joint positions are derived from the Kinect skeleton stream and converted into joint-angle features suitable for the target robot motion.
The same perception and learning framework is used across tasks; however, the trajectory model is trained separately for each manipulation objective.

\subsection{Feature Extraction}
\label{sec:feature_extraction}

The feature vector at each timestep combines human arm configuration, prompt-detected object positions, task state, and progress information.
Human joint localization is obtained from the Kinect Skeleton API, with all joint positions expressed in the Kinect camera coordinate frame.
Building on prior NAO motion-transfer work~\cite{r25}, the measured human arm geometry is converted into robot-compatible joint-angle features.
Let $\mathbf{P}_{\text{shoulder}}$, $\mathbf{P}_{\text{elbow}}$, and $\mathbf{P}_{\text{wrist}}$ denote the measured three-dimensional shoulder, elbow, and wrist positions.
Shoulder angles are computed first because they define the local frame used for elbow-angle estimation:
\begin{align}
\mathbf{v}_{\text{se}} &=
\frac{\mathbf{P}_{\text{elbow}}-\mathbf{P}_{\text{shoulder}}}
{\|\mathbf{P}_{\text{elbow}}-\mathbf{P}_{\text{shoulder}}\|}, \nonumber\\
\theta_{\text{roll}} &= \arctan2(\mathbf{v}_{\text{se}}\!\cdot\!\mathbf{e}_z,
\mathbf{v}_{\text{se}}\!\cdot\!\mathbf{e}_x), \nonumber\\
\theta_{\text{pitch}} &= \arcsin(\mathbf{v}_{\text{se}}\!\cdot\!\mathbf{e}_y).
\label{eq:shoulder_angles}
\end{align}
The elbow and wrist positions are then transformed by the inverse shoulder rotation, $\mathbf{R}_{\text{shoulder}}^{-1}$, before computing the normalized elbow-to-wrist vector.
Elbow yaw and elbow roll in the NAO joint convention are estimated from this transformed direction, yielding target values for \textit{LeftElbowYaw}, \textit{LeftElbowRoll}, \textit{LeftShoulderRoll}, and \textit{LeftShoulderPitch}.

For manipulation tasks, YOLO-World detections provide object locations in image space, which are combined with Kinect depth data to estimate three-dimensional positions of the ball, box, or block.
When an object is temporarily not detected or exits the camera field of view, the last reliable three-dimensional estimate is retained to preserve a continuous state vector.
This design keeps the learning problem low-dimensional while retaining the information needed to adjust motion according to object placement, and it allows object categories to be changed through prompts without collecting new labels for a task-specific detector.
All input and output variables are normalized before training so that object coordinates and joint angles have comparable numerical scales and remain within NAO joint limits during execution.

\subsection{Temporal Model}
\label{sec:model}

\begin{figure*}[t]
\centering
\includegraphics[width=\linewidth]{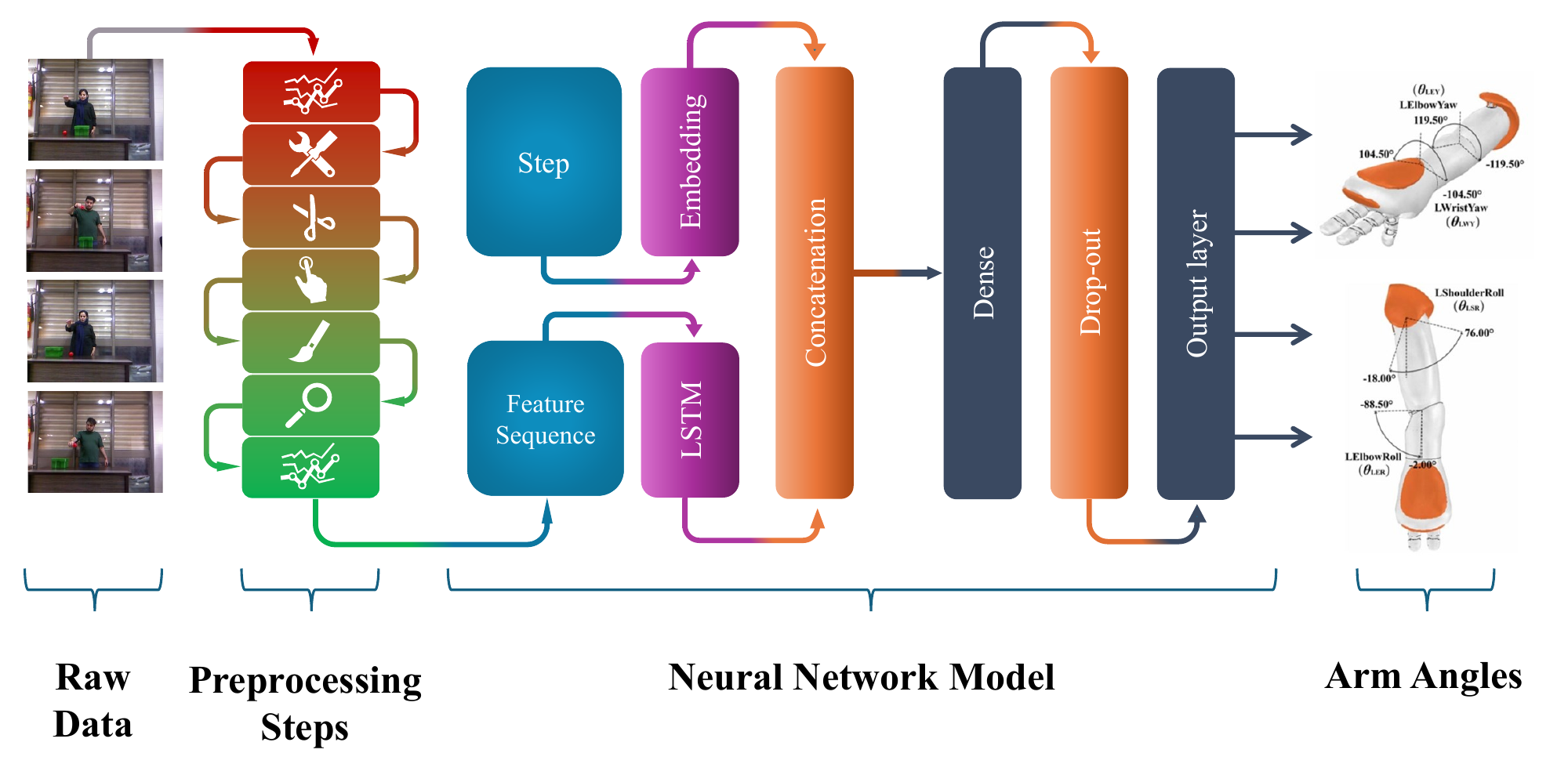}
\caption{Temporal model used to predict NAO arm joint angles from extracted demonstration features and auxiliary task embeddings.}
\label{fig:model_architecture}
\end{figure*}

The temporal model predicts robot joint angles from a fixed-length sequence of feature vectors.
As depicted in Fig.~\ref{fig:model_architecture}, a Long Short-Term Memory (LSTM) layer captures short-term motion dependencies over an input window of 10 timesteps.
Two supplementary embedding layers encode the current execution step and a three-valued grasp/task state representing the manipulation phase.
The LSTM output and embeddings are concatenated and passed through a fully connected output layer with linear activation to predict continuous robot joint angles.

The explicit step embedding provides a progress indicator during autoregressive rollout.
During deployment, the model repeatedly predicts the next joint configuration and incorporates the resulting state into the following input window.
Without such a signal, recurrent models trained on limited datasets may remain near familiar states or cycle through incomplete motions.
The step index provides a clear indication of execution progress, while the LSTM focuses on the relationship among current features, object positions, and joint trajectories.

\subsubsection{Sequence Handling and Temporal Windowing}
Variable-length demonstrations are processed at their original temporal resolution rather than being standardized to a fixed duration.
Each network input consists of a fixed window of 10 feature vectors, with the $k$-th element sampled at time $t-k\,s$ for $k\in\{0,\ldots,9\}$ and step size $s$.
Sequences shorter than the specified window are zero-padded at the beginning to support prediction from the initial frames.
During training, step sizes ranging from 5 to 95 frames are used, producing approximately $(L-1)\times10$ windows for a demonstration of length $L$, thereby exposing the model to different demonstration speeds.
During deployment, the step size controls execution speed, the step embedding is incremented with each prediction, and a conservative limit of 200 prediction steps is imposed to prevent unbounded rollout.

Model selection involved evaluating 153 configurations with different layer counts, hidden sizes, learning rates, 
batch sizes, dropout rates, and optimizers. 
Validation loss was used as the primary offline selection criterion because it provides a consistent measure 
before robot execution and allows many configurations to be compared efficiently. 
However, task-level success is not guaranteed to be perfectly correlated with prediction error, especially in physical robot execution. 
Therefore, the selected model is further evaluated through task execution in simulation and representative physical trials. 
Training uses the AdamW optimizer, mean squared error loss, early stopping, and learning-rate reduction. 
The final training setup is summarized in Table~\ref{tab:hyperparameters}.

\begin{table}[t]
\caption{Final temporal model configuration.}
\label{tab:hyperparameters}
\centering
\begin{tabular}{ll}
\toprule
\textbf{Parameter} & \textbf{Value} \\
\midrule
LSTM layers & 1 \\
LSTM hidden units & 256 \\
Input window & 10 timesteps \\
Step embedding & 64 dimensions \\
State embedding & 64 dimensions \\
Output joints & 4 \\
Optimizer & AdamW \\
Initial learning rate & 0.001 \\
Batch size & 4096 \\
Loss function & MSE \\
Maximum epochs & 3000 \\
Early stopping patience & 50 epochs \\
LR reduction factor & 0.2 \\
Minimum learning rate & $10^{-7}$ \\
\bottomrule
\end{tabular}
\end{table}

\subsection{Human-to-Robot Execution}
\label{sec:execution}

The learned commands are executed on a NAO humanoid robot.
Because humans and NAO differ in size, reachability, and joint limits, the task setup is configured so that the demonstrated object positions remain within the robot workspace.
Predicted joint commands are constrained to the feasible robot range before execution.
In simulation, object and joint positions are measured relative to a reference plane aligned with the Kinect coordinate system, ensuring that simulated inputs remain compatible with the real-world preprocessing pipeline.
This workspace alignment avoids the need for online replanning while preserving the relative task geometry observed in the demonstrations.
For pick-and-place, a magnet is attached to the robot hand and a small metal component is secured to the ball, allowing the experiment to focus on trajectory learning rather than dexterous grasping.
For pushing, the model predicts the arm motion required to move a block in the specified direction.

\section{Experiments}
\label{sec:experiments}

The framework is evaluated in simulation and on a physical NAO robot.
The experiments examine whether the proposed feature-level model can support humanoid imitation and manipulation from limited demonstrations.
Three scenarios are considered: upper-body imitation, pick-and-place, and object pushing.
In the manipulation experiments, a trial is considered successful when the robot completes the task objective without leaving the feasible workspace: 
placing the ball in the box for pick-and-place, or moving the block in the specified direction for pushing.
\subsection{Upper-Body Imitation}
\label{sec:upper_body}

\begin{figure}[t]
\centering
\includegraphics[width=\linewidth]{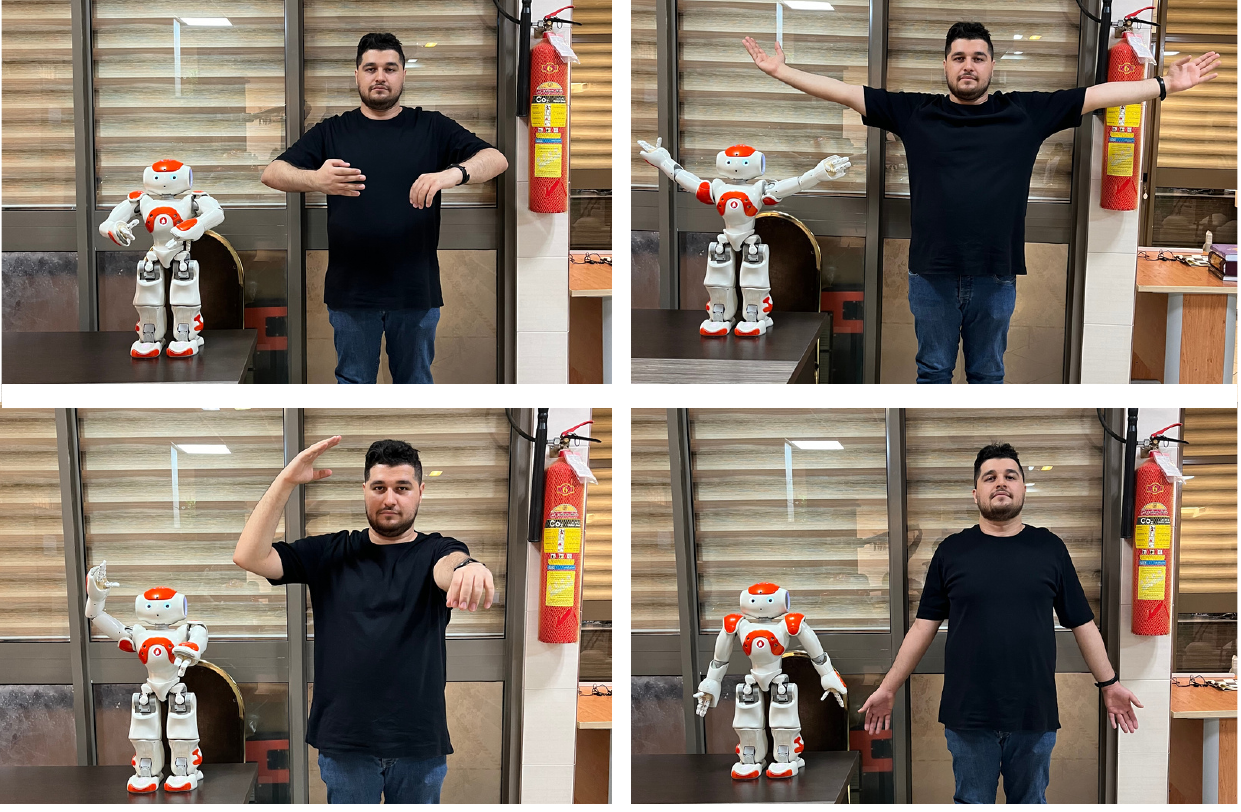}
\caption{Representative upper-body imitation results showing NAO arm configurations generated from human demonstrations.}
\label{fig:upper_body_imitation}
\end{figure}

The upper-body imitation experiment confirms that Kinect-based joint localization and human-to-robot angle mapping, 
building on previous NAO motion-transfer research~\cite{r25}, can produce feasible NAO arm motion.
Human subjects perform arm movements in front of the Kinect camera, and the derived joint angles are transformed into NAO commands.
Preliminary assessments are performed in Choregraphe to ensure motion safety before executing the commands on physical hardware.
As illustrated in Fig.~\ref{fig:upper_body_imitation}, the robot reproduces representative arm configurations through coordinated shoulder and elbow movements.
This experiment validates that the feature extraction process provides a viable basis for the manipulation studies.

\subsection{Pick-and-Place}
\label{sec:pick_place}

\begin{figure}[t]
\centering
\includegraphics[width=0.82\linewidth]{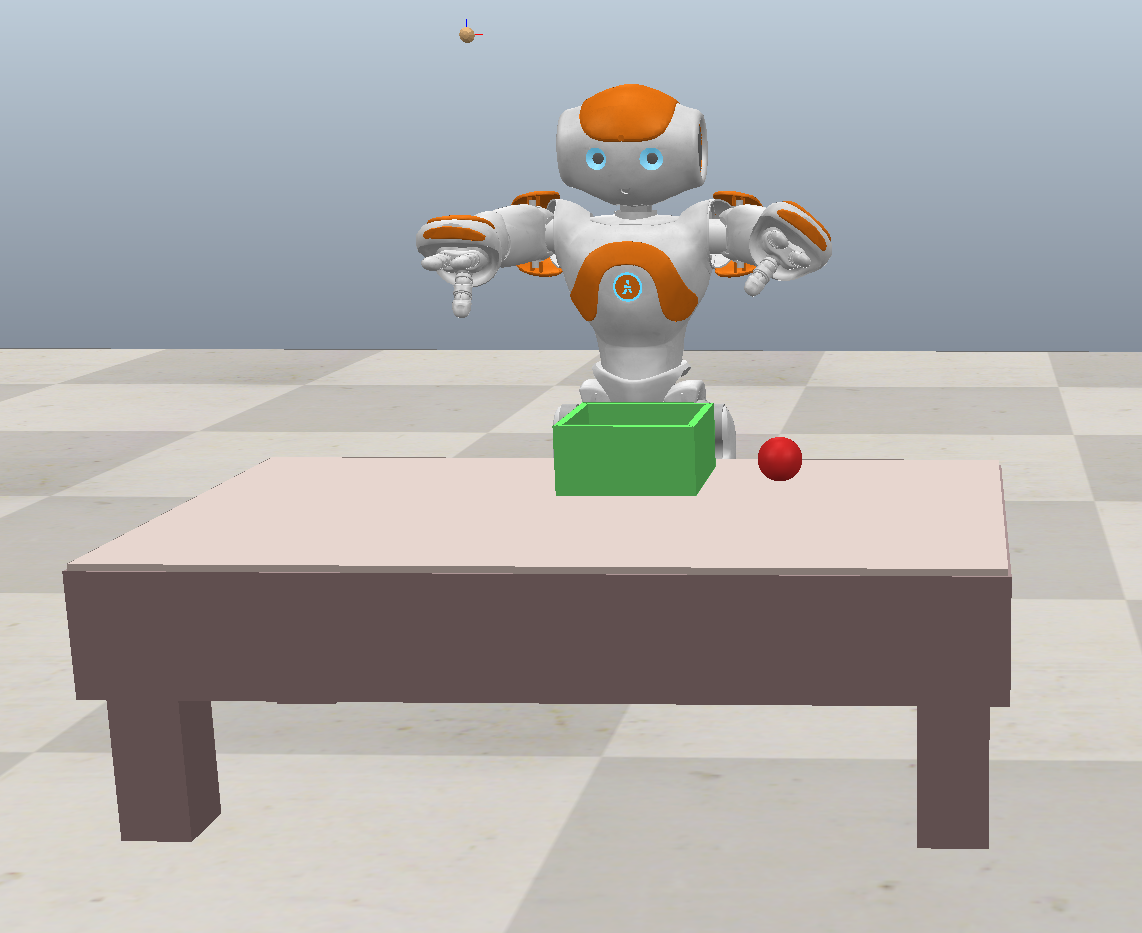}
\caption{CoppeliaSim configuration for the pick-and-place task, with the NAO robot moving a red ball into a green box.}
\label{fig:simulation_setup}
\end{figure}

The primary manipulation task is to place a red ball into a green box.
Demonstrations are recorded with variations in the initial positions of the ball and box.
The initial low-data setting contains 16 videos collected from two participants.
For comparison with larger-data settings, an additional dataset of 100 demonstrations is collected with varied object placements within the robot workspace.
The same preprocessing and model architecture are used in both settings.
The 16-video dataset was gathered in two stages, with eight initial demonstrations followed by eight additional demonstrations using different object locations.
The CoppeliaSim evaluation environment is shown in Fig.~\ref{fig:simulation_setup}.

In simulation, the proposed framework achieves a 78\% success rate with only 16 demonstrations and an 86\% success rate with 100 demonstrations.
Table~\ref{tab:pick_place} presents these results alongside relevant baselines from a related pick-and-place benchmark in Transporter Networks~\cite{r13}.
Because the tasks, robots, and sensing assumptions are different, this comparison is intended as a contextual data-efficiency reference rather than a controlled head-to-head benchmark.
The comparison supports the data-efficiency motivation for using compact object and motion features rather than learning a policy directly from raw images.

Physical robot trials are performed with the box positioned on both sides of the ball and with distracting objects visible in the camera field of view.
These trials are used to verify real-world feasibility of the perception and execution pipeline rather than to provide a full statistical sim-to-real benchmark.
The robot successfully completes representative physical executions in both left-side and right-side placement configurations.
Accordingly, quantitative success rates are reported for the controlled simulation experiments, while the physical trials demonstrate that the learned trajectories and prompt-based localization can be transferred to the NAO hardware in the tested settings.

\subsection{Pushing Task}
\label{sec:pushing}

The framework is also evaluated on a pushing task to examine its applicability beyond pick-and-place.
The robot must push a block either from right to left or from left to right.
A total of eight demonstrations are collected, with four for each direction.
The same preprocessing pipeline is used, with the block position included as an object feature.
Two direction-specific models are developed because, unlike the pick-and-place task where the box defines the target, the pushing objective is determined by the specified direction.

In the right-to-left configuration, the robot succeeds in 37 out of 50 trials.
In the left-to-right configuration, it succeeds in 34 out of 50 trials.
Overall, the pushing task achieves a 71\% success rate using only eight demonstrations.
This result indicates that the framework can be applied to another simple manipulation behavior by collecting a small task-specific dataset and retraining the temporal model.

\section{Discussion}
\label{sec:discussion}

The findings suggest that feature-level perception can support humanoid manipulation learning with limited demonstration resources.
Rather than learning both perception and control from raw images, the framework uses existing perception modules to extract human joint positions, object locations, and task state.
This compact state representation reduces the complexity of the temporal learning problem and supports adaptation across the tested variations in object placement.

\begin{table}[t]
    \caption{Contextual pick-and-place success rate comparison.}
    \label{tab:pick_place}
    \centering
    \resizebox{\columnwidth}{!}{%
    \begin{tabular}{lcccc}
    \toprule
    \textbf{Method} & \textbf{1 demo.} & \textbf{$<$20 demos.} & \textbf{100 demos.} & \textbf{1000 demos.} \\
    \midrule
    Conv. MLP~\cite{r13} & 0.0 & 3.0 & 25.5 & 31.3 \\
    GT-State MLP~\cite{r13} & 0.0 & 0.0 & 3.0 & 82.2 \\
    2-step GT-State MLP~\cite{r13} & 0.0 & 0.0 & 19.0 & 92.8 \\
    Proposed framework & 0.0 & 78.0 & 86.0 & -- \\
    \bottomrule
    \end{tabular}
    }
    \end{table}

This method also has limitations.
Kinect skeleton tracking may be inaccurate under occlusion or unusual arm positions, and detection failures can reduce the accuracy of the object state estimate.
The quantitative pick-and-place comparison is simulation-based and contextual, while the physical robot trials demonstrate feasibility rather than broad robustness or a complete sim-to-real success-rate analysis.

The current pick-and-place setup uses a magnetic attachment between the NAO hand and the ball.
This design reduces grasping uncertainty and allows the experiments to focus on trajectory learning, object localization, and humanoid arm execution.
Therefore, the reported pick-and-place results should not be interpreted as a full evaluation of dexterous grasping or general-purpose humanoid manipulation.
Removing the magnetic attachment would likely introduce additional grasping failures and requires a separate evaluation with a more realistic end-effector.

The human-to-NAO motion mapping is also approximate because the human arm and the NAO arm have different kinematic structures, joint limits, and wrist capabilities.
In particular, shoulder and elbow rotations may not map uniquely to the NAO 4-DOF arm, which can introduce end-effector pose uncertainty.
In addition, validation loss was used as the main offline model-selection criterion, although task success may not always vary monotonically with prediction error.
Future work should extend the framework to a wider range of manipulation skills, include task-level success in model selection, improve closed-loop correction during execution, and evaluate the method using more advanced sensors and dexterous end-effectors.

\section{Conclusion}
\label{sec:conclusion}

This paper introduced a data-efficient LfD framework for humanoid robots based on feature-level perception and prompt-based object detection.
The framework extracts human motion and object-location features from demonstration videos, trains a temporal model to predict NAO joint trajectories, evaluates the learned skills in simulation, and demonstrates them on physical hardware.
Experiments show successful upper-body imitation, simulated pick-and-place success rates of 78\% using 16 demonstrations and 86\% using 100 demonstrations, and a 71\% success rate in pushing with eight demonstrations.
These results demonstrate that compact task-relevant features can support the evaluated humanoid skills under constrained demonstration budgets.

\bibliographystyle{IEEEtran}
\bibliography{Bibliography}

\end{document}